\documentclass[letterpaper, 10 pt, conference]{ieeeconf}  

\IEEEoverridecommandlockouts                              

\usepackage{graphics} 
\usepackage{epsfig} 
\usepackage{times} 
\usepackage{amsmath} 
\usepackage{amssymb}  
\usepackage{pifont} 
\usepackage{multicol}
\usepackage{multirow}
\usepackage{makecell}
\usepackage[bookmarks=true]{hyperref}
\usepackage{booktabs}
\usepackage{xcolor}
\usepackage{cite} 
\usepackage{duckuments} 
\usepackage{subcaption}
\usepackage{cuted}
\usepackage{slashbox}
\usepackage{float}

\newcommand{\cmark}{\ding{51}}
\newcommand{\xmark}{\ding{55}}

\title{\LARGE \bf
MultiGripperGrasp: A Dataset for Robotic Grasping \\ from Parallel Jaw Grippers to Dexterous Hands
}

\author{Luis Felipe Casas$^{*}$, Ninad Khargonkar$^{*}$, Balakrishnan Prabhakaran, Yu Xiang%
\thanks{$^{*}$ Equal contribution. All authors are with Department of Computer Science, University of Texas at Dallas, Richardson, TX 75080, USA {\tt\small \{luis.casasmurillo, ninadarun.khargonkar, bprabhakaran, yu.xiang\}@utdallas.edu}}
}

\begin{document}

\maketitle
\thispagestyle{empty}
\pagestyle{empty}

\begin{strip}
\begin{minipage}{\textwidth}\centering
\vspace{-30pt}
\includegraphics[width=\textwidth]{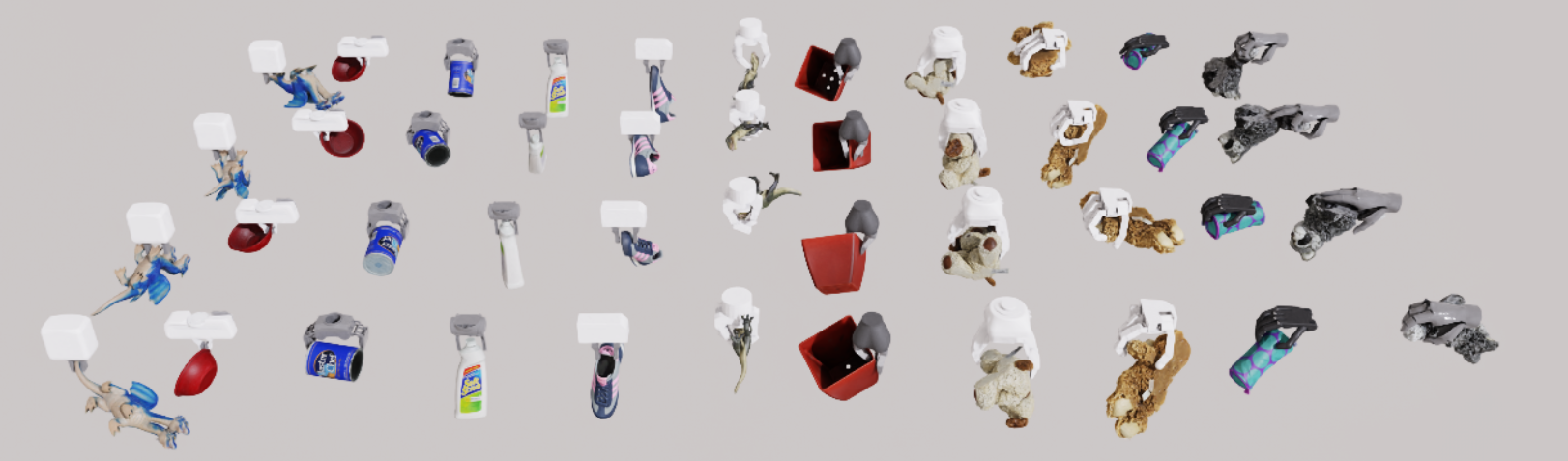}
\captionof{figure}{Dataset Visualization: Our dataset contains grasps from 11 different grippers on a diverse object set.}
\label{fig:intro-dataset-viz}
\end{minipage}
\end{strip}

\begin{abstract}

We introduce a large-scale dataset named MultiGripperGrasp for robotic grasping. Our dataset contains 30.4M grasps from 11 grippers for 345 objects. These grippers range from two-finger grippers to five-finger grippers, including a human hand. All grasps in the dataset are verified in the robot simulator Isaac Sim to classify them as successful and unsuccessful grasps. Additionally, the object fall-off time for each grasp is recorded as a grasp quality measurement. Furthermore, the grippers in our dataset are aligned according to the orientation and position of their palms, allowing us to transfer grasps from one gripper to another. The grasp transfer significantly increases the number of successful grasps for each gripper in the dataset. Our dataset is useful to study generalized grasp planning and grasp transfer across different grippers. \footnote{Data, code and videos for the project are available at \\ \url{https://irvlutd.github.io/MultiGripperGrasp}}

\end{abstract}

\section{INTRODUCTION}

Developing generalized and robust grasping methods for object manipulation is a core research topic in robotics. Autonomous robot manipulation requires grasping methods that can successfully grasp the desired object for any given task. An important problem in robotic grasping is grasp planning or grasp synthesis, where the focus is on generating the configuration of a robot gripper for grasping based on the perceived object information.

Traditional grasp planning methods use grasp quality measurements, such as task wrench space analysis~\cite{ferrari1992planning,borst2004grasp} or force closure analysis~\cite{nguyen1988constructing,berenson2008grasp} to analytically synthesize grasps. The main limitation of these approaches is that they require full-state information of the target object, which is usually represented by a 3D mesh for grasp planning. Consequently, these methods cannot synthesize grasps from partial observations, such as images or point clouds of objects. Therefore, learning-based grasp synthesis approaches~\cite{mousavian2019-6dofgraspnet,liu2020deep-ddg,sundermeyer2021-contact-graspnet,li2023gendexgrasp,attarian2023geometry} have attracted more attention in recent years. These methods use machine learning models, especially deep neural networks, to predict grasps from sensory input. By training on data with diverse objects, these methods can synthesize grasps of novel objects.

However, learning-based grasp planning requires large-scale datasets for training. Therefore, several efforts have been devoted to building grasping datasets for learning. For example, the ACRONYM dataset~\cite{eppner2021acronym} used 8,872 3D objects and generated 17.7M parallel-jaw grasps of the Franka panda gripper based on physics simulation. The DexGraspNet dataset~\cite{wang2023dexgraspnet} used 5,355 3D objects to generate 1.32M grasps for the Shadow hand. We notice that most grasping datasets are constructed for one specific gripper. As a result, the models trained on these datasets cannot be easily transferred to different grippers. A large-scale grasping dataset with a diverse set of grippers can be useful to study generalized grasp planning among various grippers.

\begin{table*}[!ht]
\centering
\caption{Overview of representative grasping datasets. +ve and -ve indicate binary successful and failed grasps, respectively.}
\label{tab:related-works-comparison}
\resizebox{\linewidth}{!}{
    \begin{tabular}{@{}cccccccc@{}}
    \toprule
    Dataset          & Year                & \#Grippers      & \#Objects & \#Grasps        & Grasp Label   & Synthesis Method        & Code + Data\\ \midrule
    HO-3D~\cite{hampali2020honnotate-ho3d} & 2020   & 1 (Human hand)             & 10                  & 78K                   & Only +ve               & Human Demo                      & \textcolor{green}{\cmark} \\

    EGAD~\cite{morrison2020egad}      & 2020        & 1 (2-finger)     & 2,331                & 233K                  & Only +ve               & Evolutionary Algorithm    & \textcolor{green}{\cmark} \\

    DDG~\cite{liu2020deep-ddg}     & 2020           & 1 (5-finger)           & 500                 & 50K                   & Only +ve   & GraspIt + modified Q1~\cite{ferrari1992planning-Q1} & \textcolor{red}{\xmark} \\

    DexYCB~\cite{chao2021dexycb}       & 2021       & 1 (Human hand)             & 20                  & 582K                  & Only +ve               & Human Demo                      & \textcolor{green}{\cmark}       \\
    
    Acronym ~\cite{eppner2021acronym}  & 2021       & 1 (2-finger)     & 8,872                & 17.7M                 & +ve \& -ve & Flex~\cite{macklin2014unified-flex-simulator} & \textcolor{green}{\cmark} \\
    UniGrasp ~\cite{shao2020unigrasp}  & 2020       & 12 (2 \& 3finger)     & 1000                &  2M+                   & Only +ve          & Contact Points Network + FastGrasp~\cite{pokorny2013classical-fastgrasp}                                 & \textcolor{green}{\cmark} \\
    
    DexGraspNet~\cite{wang2023dexgraspnet}  & 2023  & 1 (5-finger)           & 5,355                & 1.3M                  & Only +ve               & Differentiable grasping  & \textcolor{green}{\cmark} \\
    Fast-Grasp'D~\cite{turpin2023fast-graspd} & 2023 & 3 (3-5 finger)       & 2,350                & 1M                    & Only +ve               & Differentiable grasping  & \textcolor{red}{\xmark} \\
    GenDexGrasp~\cite{li2023gendexgrasp}  & 2023    & 5 (2-5 finger)       & 58                  & 436K                  & Only +ve               & Differentiable grasping  & \textcolor{green}{\cmark} \\
    \midrule
    \textbf{MultiGripperGrasp (Ours)}       & \textbf{2024}               & \textbf{11 (2-5 finger \& Human)} & \textbf{345}        & \textbf{30.4M}         & \textbf{Ranked} & \textbf{GraspIt + Isaac Sim~\cite{nvidia2023-isaac-sim}} & \textcolor{green}{\cmark}      \\ \bottomrule
\end{tabular}
}

\end{table*}

In this work, we introduce MultiGripperGrasp, a large-scale grasping dataset for studying generalized robotic grasping. Our dataset contains 30.4M grasps generated using 345 3D objects and 11 robotic grippers. There are 5 two-finger grippers, 3 three-finger grippers, 1 four-finger gripper, and 2 five-finger grippers in the dataset. One of the five-finger grippers is a human hand. Fig.~\ref{fig:intro-dataset-viz} shows examples of grasps from our dataset. Compared to existing grasping datasets, our dataset has the following unique features. 1) It has the maximum number of grippers up to date. Only a few grasping datasets contain multiple grippers in the literature. Fast-Grasp'D~\cite{turpin2023fast-graspd} and GenDexGrasp~\cite{li2023gendexgrasp} used 3 and 5 grippers, respectively. Our dataset has more diverse grippers. 2) The grasps in our dataset are ranked according to the object fall-off time in Isaac Sim~\cite{nvidia2023-isaac-sim}. In order to verify whether a grasp is successful or not, we close the fingers of a gripper according to the grasp in Isaac Sim and check if the object will fall or not. The fall-off time of the grasped object is a good measurement of grasp quality. We recorded the fall-off time for each grasp in our dataset, and used it to rank the grasps. Ranking of grasps is useful for learning models that consider grasp quality in grasp synthesis. 3) The grippers in our dataset including a human hand, are aligned using a fixed grasp pose convention for the palm. This property enables us to transfer grasps from one gripper to another gripper. The grasp transfer significantly increases the number of successful grasps for each gripper in our dataset. Since grasp transfer is critical in skill sharing between robots and skill learning from human demonstration, our dataset can be used to study grasp transfer between different grippers.


In summary, our paper has the following contributions:
\begin{itemize}

    \item We constructed a large-scale robotic grasping dataset that contains 30.4M grasps of 11 different grippers on 345 objects. These grippers range from two-finger grippers to five-finger grippers, including a human hand.
    
    \item The grasps in our dataset are verified in Isaac Sim, and ranked according to the object fall-off time. The grasp ranking can be used to learn models for grasp quality prediction.

    \item The grippers in our dataset are aligned, which can be utilized to study grasp transfer among different grippers.

\end{itemize}

\section{RELATED WORK}

\subsection{Grasping Datasets}

Learning-based grasp planning methods are crucially dependent on a labeled dataset of high-quality grasps for training. Table~\ref{tab:related-works-comparison} compares our MultiGripperGrasp dataset with existing datasets for robotic grasping.
Several previous datasets for parallel jaw grippers have been introduced for image-based grasp prediction~\cite{lenz2015deep-cornell-grasp}, with extensions to suction grippers~\cite{mahler2017dexnet-2}. Recent datasets for parallel jaw grippers~\cite{morrison2020egad,eppner2021acronym} are much larger in scale in terms of the number of objects and grasps.


Most multifinger datasets focus on a single gripper such as the Barrett hand~\cite{goldfeder2009-columbia-grasp-database, lundell2021multi-fingan, lundell2021ddgc}, the Shadow hand~\cite{liu2020deep-ddg,wang2023dexgraspnet} or human hand~\cite{hampali2020honnotate-ho3d,chao2021dexycb}.  
As a result, the models learned from these datasets cannot be easily transferred to other grippers. Only a few datasets~\cite{shao2020unigrasp,turpin2023fast-graspd,li2023gendexgrasp, wang2023dexgraspnet} provide grasps across multiple different grippers. Unigrasp~\cite{shao2020unigrasp} contains 12 different grippers in their dataset, but all have no more than 3 fingers, thereby making it insufficient for dexterous grasping. 
Several differentiable grasping methods~\cite{li2023gendexgrasp,turpin2023fast-graspd,wang2023dexgraspnet} are also used to construct grasping datasets with multiple grippers. However, these datasets are limited to 3 or 5 grippers as shown in Table~\ref{tab:related-works-comparison}.

\subsection{Grasp Synthesis Approaches}

\textbf{Analytical Methods.} Generating high-quality grasps synthetically has been studied for decades. Classical grasp synthesis methods~\cite{miller2004graspit,goldfeder2009-columbia-grasp-database,mahler2017dexnet-2} have proposed a variety of analytic solutions~\cite{ferrari1992planning-Q1,rimon1996force-closure-classical} that reason about contact wrenches between a gripper and an object.
These methods optimize a grasp quality metric~\cite{rimon1996force-closure-classical} that measures the force closure property of a grasp and its ability to resist external forces and torques. Gradient-free methods, such as simulated annealing~\cite{miller2004graspit} are used to find grasp poses and joint angles that can achieve good grasp quality.
Recent analytic methods focus on improving grasp synthesis for multi-fingered grippers. For example, \cite{ciocarlie2007dexterous-eigengrasps} constrains the optimization over a lower-dimensional search space via the concept of eigen-grasps. \cite{tsuji2009easy-friction-cone-approx} speeds up the optimization by introducing ellipsoids for friction cone approximations.

\textbf{Differentiable Grasping.} The other spectrum of grasp synthesis approaches leverages gradient-based techniques to optimize the search space using different notions of differentiable grasp quality metrics. 
Some reformulate the analytic force closure estimation in a differentiable setting, with~\cite{liu2021synthesizing-dfc} using a simplified version to increase the synthesis speed and~\cite{liu2020deep-ddg} focusing on modified contact constraints between the gripper and the object. However, these methods are still slow to converge, which limits their application to the creation of large-scale datasets. Subsequently, \cite{wang2023dexgraspnet} proposes alternatives for pose initialization and contact penetration energies.
In a similar vein, \cite{turpin2022-graspd} introduces a simulation protocol that can scale up to a higher number of contacts for contact-rich human hand manipulation and~\cite{turpin2023fast-graspd} extends it for robotic hands with a much faster grasp synthesis algorithm.

\begin{figure*}
\centering
\begin{center} 
\includegraphics[width=\textwidth]{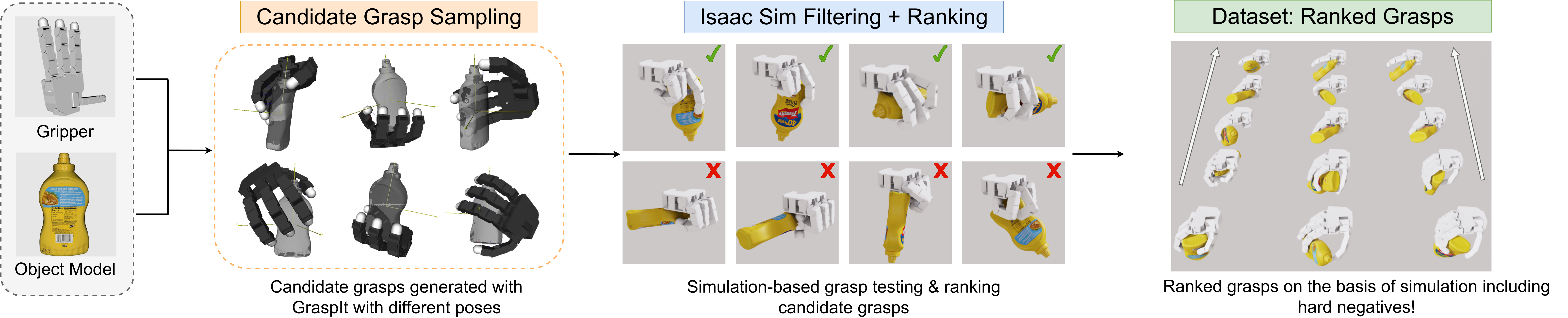}
\caption{Our dataset creation pipeline: Given a pair of an object and a gripper, GraspIt! is used to generate candidate grasps that are then filtered using Isaac Sim to produce ranked grasps.}
\label{fig:method-pipeline}
\end{center}
\end{figure*}

\begin{figure}
\centering
\begin{center} 
\includegraphics[width=\linewidth]{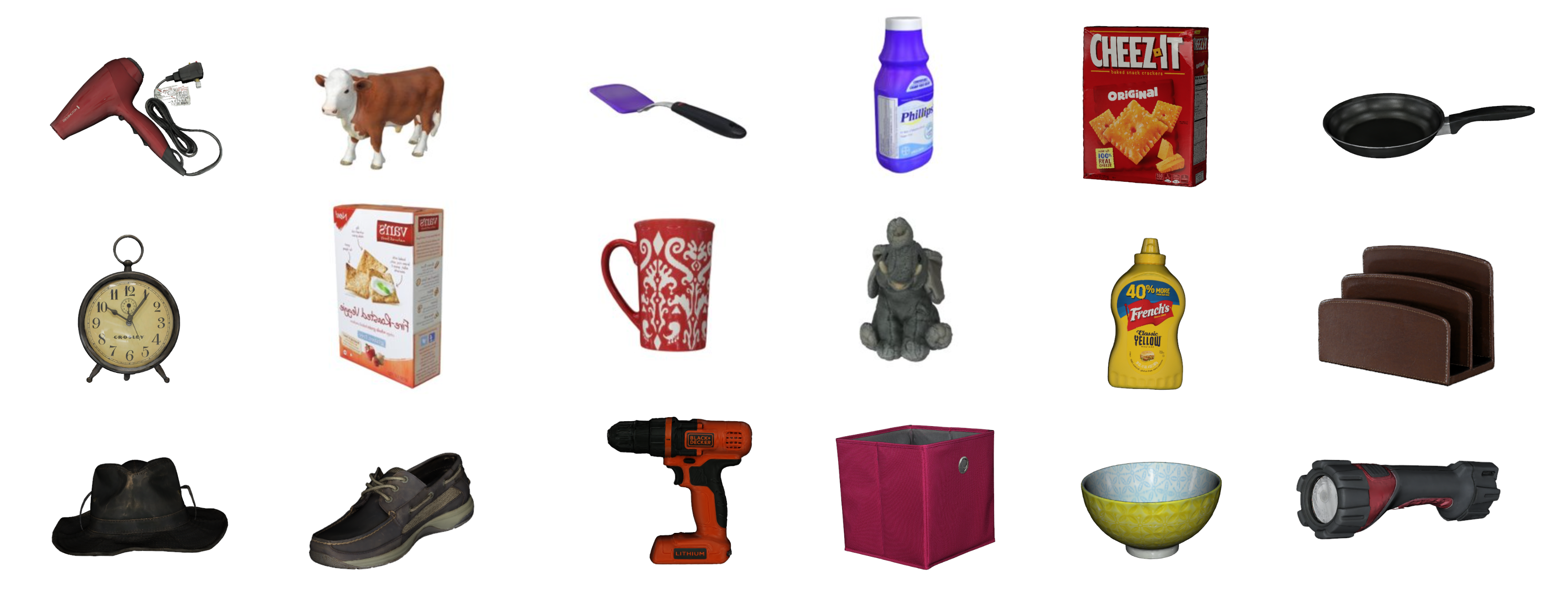}
\caption{A sample set of objects used for grasp generation.}
\label{fig:sample-objects}
\end{center}
\end{figure}

\begin{figure}
\centering
\begin{center} 
\includegraphics[width=\linewidth]{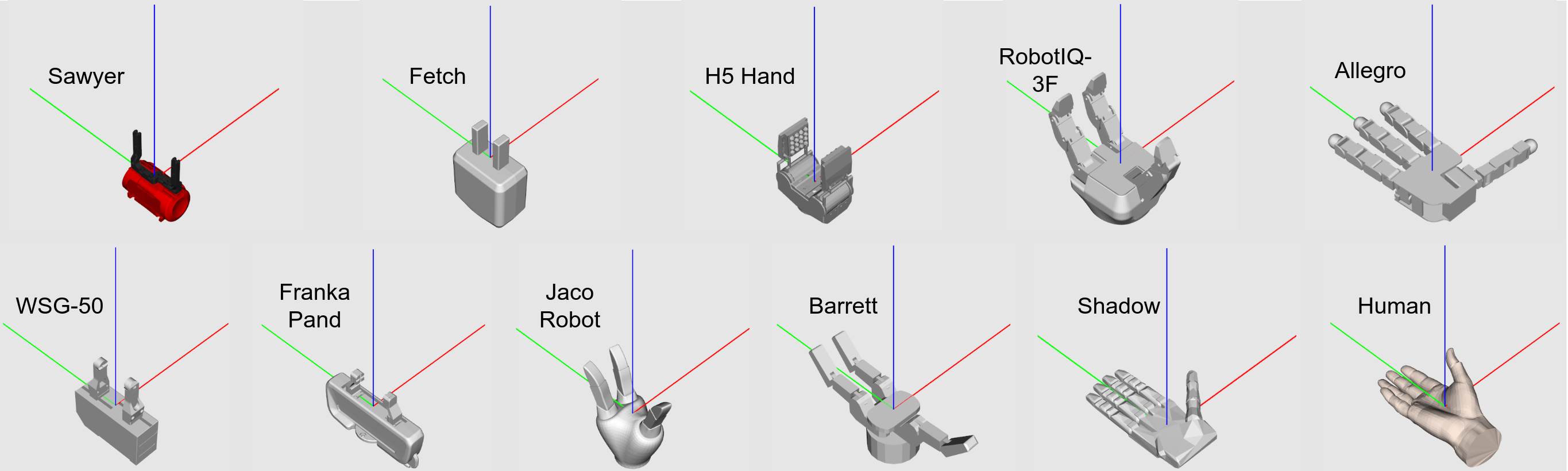}
\caption{Aligned grippers in our dataset with varying dexterity.}
\label{fig:gripper-set}
\end{center}
\end{figure}

\subsection{Learning-based Grasping}

Learning-based methods for grasping have recently found favor due to their capability of using partial observations for grasp planning, such as images or point clouds of objects. 
These methods learn direct regression mappings~\cite{mahler2017dexnet-2, lenz2015deep-cornell-grasp}, generative models~\cite{mousavian2019-6dofgraspnet, lundell2021ddgc} or implicit functions~\cite{khargonkar2023neuralgrasps, weng2023neural-ngdf} conditioned on sensory input. 
In terms of input mode, some of the earlier work used RGB-D images~\cite{mahler2017dexnet-2, lenz2015deep-cornell-grasp}. With advances in point-cloud representations~\cite{qi2017pointnet++}, recent methods such as~\cite{mousavian2019-6dofgraspnet, sundermeyer2021-contact-graspnet,weng2023neural-ngdf} leverage point clouds, making it easier to reason about contacts between grippers and objects.

In terms of gripper representation, most learning-based grasping methods have focused on parallel jaw grippers~\cite{lenz2015deep-cornell-grasp, mahler2017dexnet-2, mousavian2019-6dofgraspnet, sundermeyer2021-contact-graspnet} due to the widespread availability of large-scale datasets of parallel jaw grippers~\cite{morrison2020egad,eppner2021acronym}. 
Multifinger grasping methods usually employ generative models~\cite{lundell2021multi-fingan,lundell2021ddgc,wei2022dvgg} to reason about gripper 6D pose and finger joint values simultaneously. At inference time, a large number of grasps are sampled, which are later refined to fit better to the object. 
Several methods have been proposed to generalize the grasp prediction between different grippers by using contact points as a bridge, where contact points can be represented by implicit distance functions~\cite{khargonkar2023neuralgrasps}, auto-regressive models~\cite{attarian2023geometry} or hand-agnostic contact maps~\cite{li2023gendexgrasp}.

\section{METHOD: Dataset Generation}

The pipeline of our dataset creation is illustrated in Fig.~\ref{fig:method-pipeline}. It starts with a pair of an object and a gripper. Then a set of initial grasps is generated using GraspIt!~\cite{miller2004graspit}, followed by the filtering and ranking of these grasps in Isaac Sim~\cite{nvidia2023-isaac-sim}. We describe each component of the pipeline in detail.


\subsection{Objects and Grippers}\label{sec:method-objects-and-grippers}

\textbf{Object Set.}
We chose a subset of 329 objects from the GoogleScannedObjects dataset~\cite{downs2022google-scanned-objects} as our core object set. These are 3D scanned objects in everyday settings. The subset is chosen according to the FewSOL~\cite{padalunkal2023fewsol} dataset for few-shot object recognition in robotic settings. Furthermore, we also included 16 objects from the YCB object set~\cite{calli2015ycb-object-set} due to the popularity of the YCB objects in robot manipulation. Some examples of objects in our dataset are shown in Fig.~\ref{fig:sample-objects}.
The ground truth mass and friction coefficients of each object is unknown to us, consequently, we used a fixed density (100 $kg/m^{3}$) and friction coefficients (0.5). 



    
    


\begin{figure*} [h!]
    \centering
    \begin{center} 
    \includegraphics[width=\textwidth]{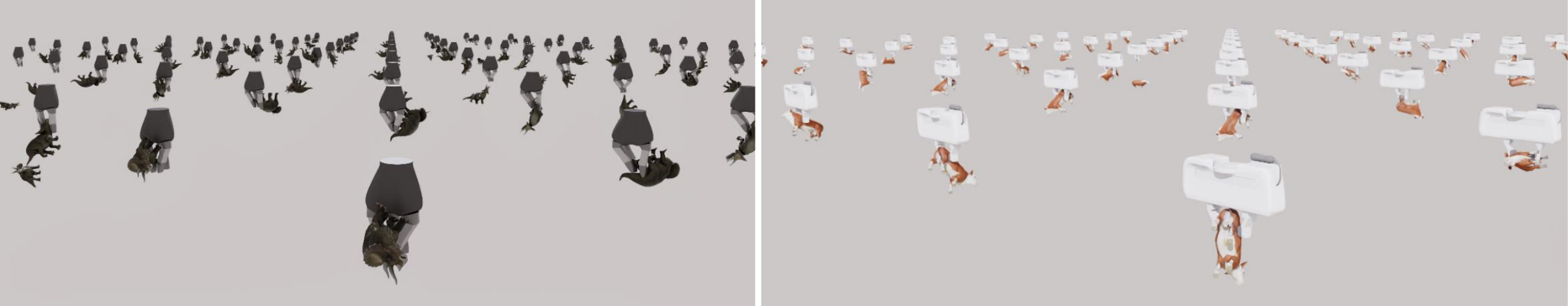}
    \caption{Illustration of the grasp ranking process with Isaac Sim.}
    \label{fig:grasp_filtering}
    \end{center}
\end{figure*}

\begin{figure*}
    \centering
    \begin{center} 
    \includegraphics[width=\textwidth]{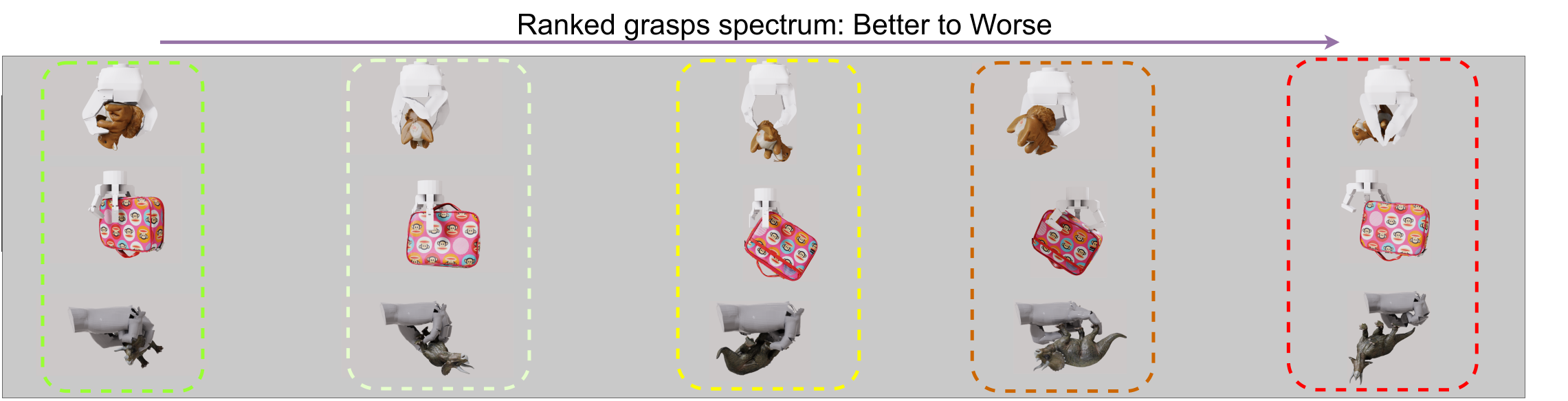}
    \caption{Visualization of ranked grasp labels and negative examples illustrating the advantage of going beyond binary labels.}
    \label{fig:ranked-grasp-labels}
    \end{center}
\end{figure*}

\textbf{Gripper Set.}
The gripper set in our dataset contains a wide variety of grippers from parallel jaw grippers to dexterous hands. 
In total, there are 11 grippers with 5 parallel jaw grippers, 3 three-finger grippers, and 3 dexterous hands as shown in Fig.~\ref{fig:gripper-set}:
\begin{itemize}
    \item 2-finger grippers: Fetch, Franka Panda, WSG50, Sawyer, H5 Hand
    \item 3-finger grippers: Barrett, Robotiq-3F, Jaco Robot
    \item 4-finger grippers: Allegro
    \item 5 finger grippers: Shadow, Human Hand
\end{itemize}
Methods using classical grasp quality metrics~\cite{miller2004graspit, liu2020deep-ddg} and contact-energy optimization require specifying contact regions on the gripper fingers and palm. For each gripper, we selected such regions using the graphical interface in Graspit. To transfer grasps between different grippers, a common notion of gripper pose was used as illustrated by the axes in Fig.~\ref{fig:gripper-set}. The translation refers to the palm center of the gripper, and the orientation is with respect to a canonical pose of the gripper palm pointing in a fixed direction. Thus, any grasp pose from a gripper can be transferred to another by using this pose alignment. For collision checking of these grippers in Isaac Sim, we use a convex decomposition over the gripper meshes.



\subsection{Grasp Generation}

We use GraspIt!~\cite{miller2004graspit} to generate candidate grasps for each pair of an object and a gripper. GraspIt! fits our needs for the initial phase of grasp generation as it is easy to integrate novel grippers into its framework and is less memory intensive.
An alternative could be to use a differentiable grasping method to generate the initial grasps. For each object-gripper pair, we generated around 8,000 candidate grasps with the caveat that not all of them will pass the simulation filtering check. Here, a contact-based energy formulation was chosen for the simulated annealing optimization used by Graspit! with manually specified contact points on all grippers.

Specifically, Graspit! proceeds by sampling a large number of 6-DOF poses for the gripper around the object as a starting configuration for optimizing a contact-based energy function.  A queue of best few grasps is updated during the optimization until a maximum number of iterations are reached, after which the grasp is executed and analytical quality metrics are computed. Having a large number of grasp proposals was necessary to ensure grasp diversity.

\subsection{Grasp Labeling}

\textbf{Motivation.}
To further evaluate the quality of the grasps generated by GraspIt!, we use a robotics simulator known as Isaac Sim~\cite{nvidia2023-isaac-sim} to verify these grasps, since GraspIt! only considers object and gripper geometry for grasp analysis. Crucially, we provide a ranked order for these grasps based on the simulation statistics instead of simply providing grasps that pass the simulation test as positive grasps. 
This allows some grasps in the dataset to be used as negative examples in learning-based methods that use grasp evaluators such as 6D GraspNet~\cite{mousavian2019-6dofgraspnet}. These negative grasps are of higher quality than random sampled grasps, since they represent hard negative examples during training.


\textbf{Procedure.} The general idea of grasp labeling is to check if a grasped object will fall or not due to gravity in simulation as illustrated in Fig.~\ref{fig:grasp_filtering}. To test the grasps in Isaac Sim, the gripper is placed so that its palm normal is aligned with the direction of gravity, and the object is initialized according to its pose with respect to the gripper. This set up allowed us to reliably make gravity a adversarial force to every grasp. The joint positions of the gripper fingers are then initialized according to the given grasp. We designed a controller to gradually close the gripper fingers until contact with the object is established. Contact between the fingers and the object triggers the gravity force. Subsequently, physics is simulated over a fixed time interval or until the object falls.  The maximum grasp test time was set to 3 seconds.



\textbf{Stability Measure.}
Grasps are evaluated according to the time stamp when the object falls. When gravity is applied, a timer is started. Then we recorded the fall-off time for the object as a grasp quality measurement, with higher values for more stable grasps. The object fall-off time is recorded either when the object loses contact with the gripper fingers or when the maximum test time has expired. In the latter case, the grasp will be given the maximum test time as the fall-off time. Furthermore, the object is considered to have fallen when the current pose is 0.5 meters away from its initial position. If the gripper cannot make contact with the object, a negative value is assigned as the fall-off time, and the grasp is considered as a failed grasp. The gripper contacts with the object were calculated using the Rigid Body API from Isaac Sim~\cite{nvidia2023-isaac-sim}. Fig.~\ref{fig:ranked-grasp-labels} shows some ranked grasps according to their object fall-off time.



\begin{figure}
    \centering
    \begin{center} 
        \includegraphics[width=\linewidth]{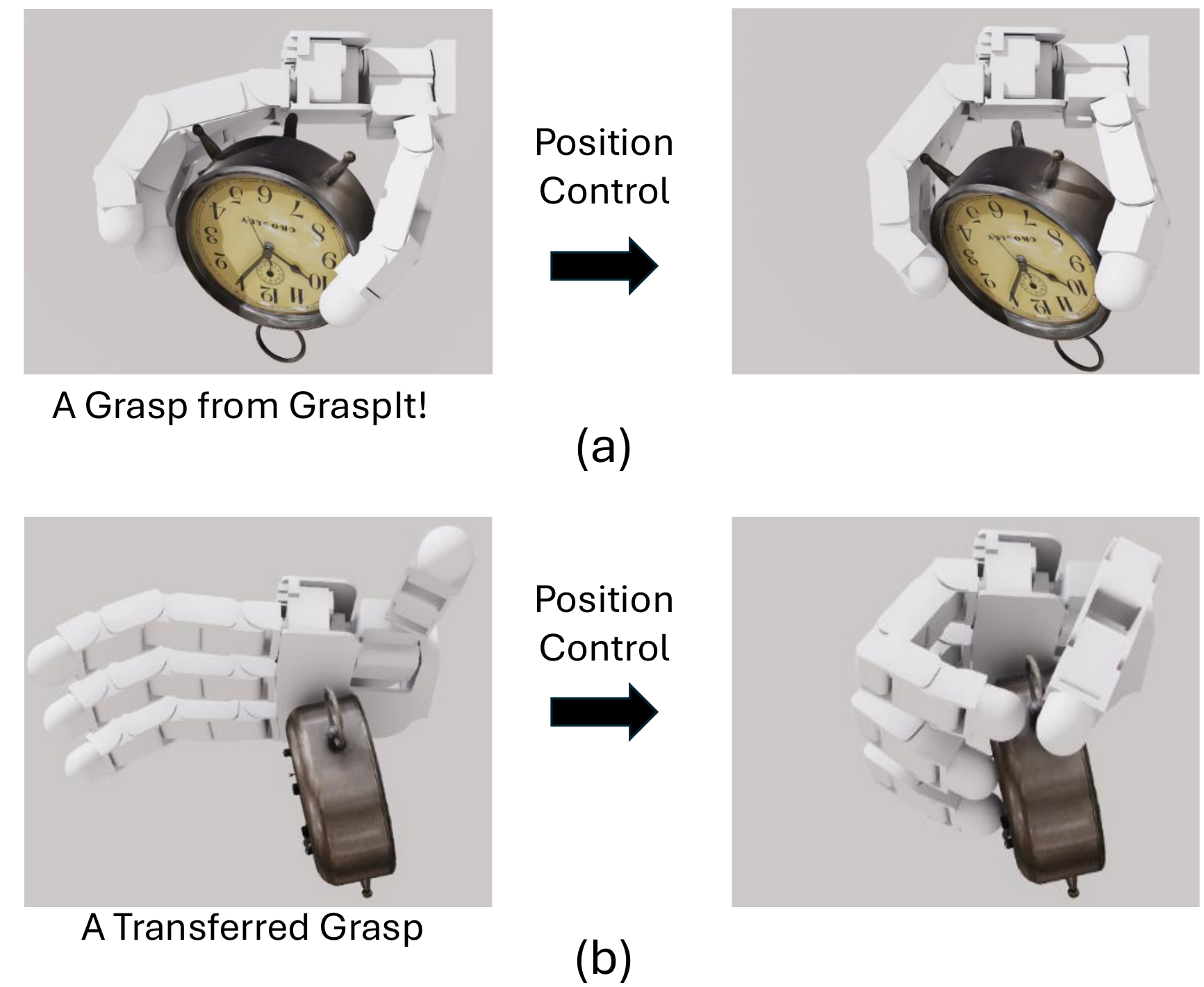}
        \caption{Position control for closing fingers. (a) A grasp generated from GraspIt! (b) A transferred grasp } 
        \label{fig:controller}
    \end{center}
    \vspace{-4mm}
\end{figure}

\subsection{Controller Design} 
\label{sec:controller}
The grasps, when initialized within Isaac Sim, are in contact or close to contact with the objects. However, a gripper controller is needed to exert force on the object to perform the grasp. For parallel jaw grippers, the control can be performed by setting the two fingers to their close position. On the other hand, for multi-finger grippers some joints must remain static, e.g., lateral movement of the fingers. Given the large amount of gripper, object, and pose combinations within our dataset, we opted for a simple controller that could be easily used with all the grippers. For each gripper, we manually identify a subset of joints to move and use a position controller to reach the gripper closing configuration. The following versions of the controller were implemented:


\textbf{Position Control for Grasp Ranking. }For grasps generated from GraspIt!, We simply control the root joints of the fingers to reach their closing positions  (see Fig.~\ref{fig:controller}(a)). The closing positions are manually specified for each gripper. 

\textbf{Position Control for Grasp Transfer.} To transfer a grasp from a source gripper to a target gripper, we first set the joint configuration of the target gripper to the open configuration. Then the controller closes all fingers until it comes into contact with the object. After that, the controller sets only the root joints of the fingers to reach their closing position (see Fig.~\ref{fig:controller}(b).

\begin{figure*}
    \centering
    \begin{center} 
        \includegraphics[width=\linewidth]{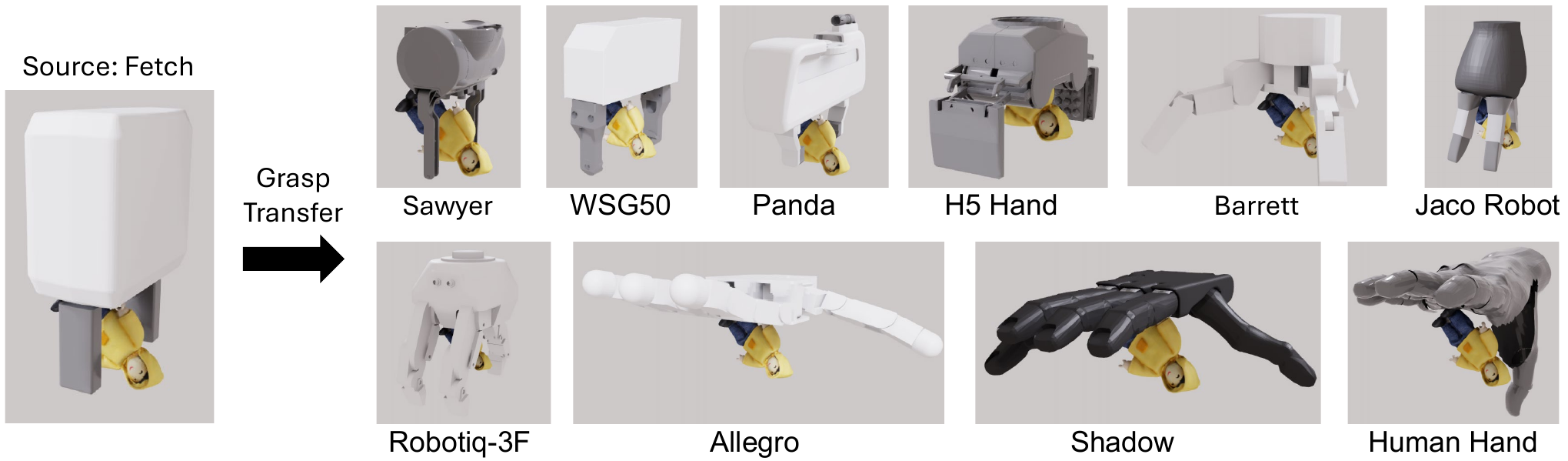}
        \caption{An example for grasp transfer: original GraspIt! grasp (left), transferred grasps (right).}
        \label{fig:transfer}
    \end{center}
\end{figure*}

\begin{table}
\centering
\caption{Success rates and failure rates against object fall-off times for different grippers in our dataset.}
\label{tab:sucess_rate}
\resizebox{\linewidth}{!}{
    \begin{tabular}{@{}cccccccc@{}}
    \toprule
    \multirow{2}{*}{Gripper}       & \multirow{2}{*}{\#Fingers} & \multicolumn{4}{c}{Success Rate (\%)}   &  Failure &  \#Grasps  \\ 
           & & $>$ 3s  & $>$ 2s   & $>$ 1s   &  $>$ 0s    & Rate (\%)   & (M)   \\ \midrule
    Fetch             &    2    &  69.42  &  69.64  &  70.37  &  96.34  &  3.66  &      2.73        \\
    Franka      &    2    &  70.17 &  70.43  &  71.05  &  98.09  &  1.91  &      2.76          \\
    H5 Hand              &    2    &  1.18   &  1.27   &  1.47   &  83.9  &  16.4 &      3.38          \\
    Sawyer            &    2    &  48.15   &  48.60  &  49.77   &  87.31  &  12.69 &      2.49           \\
    WSG 50            &    2    &  61.29  &  61.51  &  62.24  &  96.86  &  3.14  &      2.75  
    \\ 
    Barrett           &    3    &  45.29  &  46.12  &  49.39  &  93.79  &  6.21  &      2.62          \\
    Jaco         &    3    &  52.20  &  52.76  &  54.3  &  98.96  &  1.04  &     2.76          \\
    Robotiq           &    3    &  \textbf{76.09}  &  \textbf{77.03}  &  \textbf{79.03}  &  \textbf{99.94}  &  \textbf{0.06}  &      2.72          \\
    Allegro           &    4    &  44.27  &  45.76  &  50.15  &  99.87  &  0.13  &      2.77           \\
    Shadow            &    5    &  4.37   &  4.67   &  6.72   &  96.66  &  3.34  &      2.78          \\
    Human         &    5    &  13.56  &  14.38  &  17.21  &  92.84  &  7.16  &      2.65       \\
     \bottomrule
\end{tabular}
}
\end{table}

\begin{table*} [ht]
\centering
\caption{Confusion matrix of grasp transfer. Success rates of grasps transferred from a source gripper (row) to a target gripper (column) are presented. The last column represents the number of unique grasps transferred from the source gripper, and the last row is the number of successful grasps across all the transferred grasps for the target gripper.} 
\label{tab:confusion}
\resizebox{0.98\linewidth}{!}{
    \begin{tabular}{@{}ccccccccccccc@{}}
    \toprule
            \backslashbox{Source}{Target}   &  Fetch &  Franka  &  H5 Hand  &  Sawyer  &  WSG 50  &  Barret  &  Jaco & Robotiq & Allegro & Shadow & Human &   \makecell{\#Source \\ Grasps (K)}  \\ \midrule
    Fetch             &1&0.91&0.84&0.86&0.45&0.71&0.85&0.77&0.32&0.38&0.25& 147  \\
    Franka            &0.84&1&0.71&0.80&0.38&0.64&0.72&0.68&0.31&0.37&0.22 & 145 \\
    H5 Hand           &0.05&0.02&1&0.03&0.04&0.45&0.11&0.50&0.16&0.07&0.03& 24 \\
    Sawyer            &0.88&0.76&0.57&1&0.15&0.57&0.71&0.65&0.29&0.30&0.15& 117 \\
    WSG 50            &0.46&0.39&0.36&0.51&1&0.56&0.57&0.63&0.23&0.25&0.12 & 152 \\ 
    Barrett           &0.60&0.49&0.48&0.60&0.41&1&0.59&0.77&0.23&0.29&0.19 & 149 \\
    Jaco              &0.76&0.65&0.64&0.70&0.31&0.70&1&0.74&0.30&0.35&0.23 & 140 \\
    Robotiq           &0.42&0.34&0.35&0.39&0.31&0.58&0.38&1&0.19&0.23&0.13 & 167 \\
    Allegro           &0.18&0.12&0.05&0.26&0.16&0.47&0.30&0.51&1&0.42&0.13& 158 \\
    Shadow            &0.18&0.11&0.06&0.26&0.08&0.42&0.38&0.56&0.51&1&0.10& 67 \\
    Human             &0.21&0.23&0.14&0.27&0.07&0.48&0.51&0.62&0.37&0.43&1 & 141 \\\midrule
    \makecell{\#Success Target \\ Grasps (K)}  & 778 & 706 & 622 & 789 & 489 & 873 & 836 & 987 & 518 & 508 & 358 &\\
     \bottomrule
\end{tabular}
}\end{table*}

\section{EXPERIMENTS}

To build our dataset, we conducted experiments to rank grasps generated from GraspIt! and transfer grasps from one gripper to another.

\subsection{Grasp Ranking}

For the 30.4 million grasps across 3,795 different object-gripper pairs, our simulation was able to record the object fall-off time for each grasp. 
Table~\ref{tab:sucess_rate} shows the amount of grasps simulated per gripper, the percentage of object fall-off times over a threshold per gripper and the failure rate of the grasps simulated, i.e. the percentage of grasps that were not able to make contact with the object. Based on the object fall-off time, the results showed that out of the 30.4 million grasps, 13.16M are successful grasps (the object did not fall within 3s), 0.17M lasted from 2s to 3s, 0.53M lasted from 1s to 2s, 14.97M fell shortly after the gripper touched the object (0s to 1s) and 1.59M are classified as failed grasps.

From Table~\ref{tab:sucess_rate}, we can see that the Robotiq gripper achieves the highest success rate. We observed low success rates for three grippers in our dataset: the H5 Hand, the Shadow Hand, and the human hand. For the H5 hand, GraspIt! had difficulties in planning good grasps due to the presence of unsupported mimic joints. Therefore, the grasps through simulated annealing were sub-optimal. In the case of Shadow hand and human hand, we observed that our controller is sub-optimal for many grasps when considering highly articulated grippers with complex joints. We observed that the generated joint values for the thumbs made the optimal control non-constant. In some cases, the joints may be aligned in a way that our controller creates a stable grasp, while in others it applies adversarial forces to the grasp. The question of how to design a better controller for multi-finger hands remains a challenging task.



\subsection{Grasp Transfer} \label{sec:transfer}

Our simulation was able to evaluate the object fall-off time for a large amount of generated grasps. The successful grasps of one gripper can represent successful grasps in others and increase the overall amount of successful grasps in the dataset. To test this hypothesis, we implemented the grasp transfer of successful grasps from one gripper to others and evaluated the transferred grasps using Isaac Sim.

We utilize the alignment between grippers as described in Sec.~\ref{sec:method-objects-and-grippers} to transfer grasps.  We first transform a source gripper pose to its aligned pose, and then transform the aligned pose to the target gripper. Fig.~\ref{fig:transfer} illustrates an example for grasp transfer. To execute the new grasp, the target gripper is controlled using the position controller described in Sec.~\ref{sec:controller}. The joint values at the moment of contact between the gripper and object are recorded, and the object fall-off time is simulated.


Table~\ref{tab:confusion} shows the confusion matrix of grasp transfer in our experiment. To limit the runtime, we sampled 500 successful grasps per object-gripper pair, where a successful grasp is a grasp with an object fall-off time greater than three seconds. Many successful grasps of one gripper were successful grasps in other grippers when transferred. For example, for the Franka Panda gripper, 561K transferred grasps were evaluated as successful by the simulation. For the H5 Hand, where the original grasps generated from GraspIt! are poor, we were able to generate 598K new successful grasps from other grippers compared to the 24K original successful grasps. The table also shows that some grasps from similar grippers were not successful. Gripper reach and morphology are not taken into account when transferring grasps, thus some grasps may be bad when tested in grippers with different finger shape and size.


\section{LIMITATIONS}

First, our dataset is limited by the number of objects in the dataset. Scaling our dataset to larger object sets along with cluttered scenes would be better for downstream tasks, and such aspects can be continually improved upon. Second, most of the datasets in Table~\ref{tab:related-works-comparison} including ours do not consider affordance-driven grasps. Rather than relying solely on object geometry and physics simulation, the dataset can be expanded to include semantic grasps. Finally, our position controller for grasping is sub-optimal. More sophisticated controller design for grippers can also help to obtain better ranking metrics, such as using a force controller to adjust for object disturbances.


\section{CONCLUSION}

In this work, we present MultiGripperGrasp, a large-scale grasping dataset with 11 different grippers across the spectrum of two to five finger grippers. Moreover, we go beyond binary labels of success or failure for grasps and include a ranked order based on grasping performance in a physics simulation along with aligning the gripper poses for grasp transfer. Incorporating different classes of grippers also brings out interesting observations in our experiments, especially with respect to leveraging grasp transfer as a simple baseline for augmenting a gripper with lower performing grasps. With MultiGripperGrasp, we hope to help research on generalized grasping~\cite{li2023gendexgrasp, attarian2023geometry} and allow the learning of cross-functional skills between highly different gripper morphologies. 



\section*{Acknowledgement} This work was supported by the Sony Research Award Program.

\bibliographystyle{IEEEtran}
\bibliography{references}

\end{document}